\documentclass{article}

\usepackage{microtype}
\usepackage{graphicx}
\usepackage{subfigure}
\usepackage{booktabs} 
\usepackage{amsmath}
\usepackage{amsfonts}

\usepackage[accepted]{icml2020}

\usepackage{hyperref}

\icmltitlerunning{Autoregressive flow-based causal discovery and inference}

\begin{document}

\twocolumn[
\icmltitle{Autoregressive flow-based causal  discovery and inference}



\icmlsetsymbol{equal}{*}

\begin{icmlauthorlist}
\icmlauthor{Ricardo Pio Monti}{1}
\icmlauthor{Ilyes Khemakhem}{1}
\icmlauthor{Aapo Hyv\"{a}rinen}{2}
\end{icmlauthorlist}

\icmlaffiliation{1}{Gatsby Computational Neuroscience Unit, UCL}
\icmlaffiliation{2}{University of Helsinki}

\icmlcorrespondingauthor{}{ricardo.monti08@gmail.com}

\icmlkeywords{Machine Learning, ICML}

\vskip 0.3in
]




\printAffiliationsAndNotice{}  

\begin{abstract}
We posit that autoregressive flow models are well-suited to 
performing a range of causal inference tasks --- ranging from causal discovery to making
interventional and 
counterfactual predictions. 
In particular, we exploit the fact that 
autoregressive architectures 
define an 
ordering over 
variables,
analogous to a causal ordering, in order to 
propose a single flow architecture to perform all three aforementioned tasks.  
%
We first leverage the fact that flow models estimate normalized 
log-densities of data 
to derive a bivariate measure of causal direction based on likelihood ratios. 
Whilst traditional measures of causal direction often require restrictive assumptions on the nature of 
causal relationships (e.g., linearity),
the flexibility of flow models allows for arbitrary causal dependencies.
%
%
Our approach compares favorably against 
alternative methods on synthetic data 
as well as on the Cause-Effect Pairs benchmark dataset.
%
%
Subsequently, we demonstrate that the invertible nature of flows naturally allows for 
direct evaluation of both 
interventional and counterfactual predictions, which require 
marginalization and conditioning over latent variables respectively. 
We present examples  over synthetic data 
where autoregressive flows, when trained 
under the correct causal ordering, are able to make accurate interventional and counterfactual predictions. 
\end{abstract}

\section{Introduction}
\label{sec:intro}

Causal models play a fundamental role in modern scientific endeavor \citep{Spirtes2000, Pearl2009} 
with
many of the questions which drive research in science being not associational but 
rather causal in nature. 
%
To this end, the framework of structural equation models (SEMs) 
was developed 
to both encapsulate causal knowledge as well as 
answer 
interventional and counterfactual queries \citep{pearl2009causal}

At a fundamental level, SEMs define a generative
model for data based on 
causal relationships. 
As such, SEMs 
implicitly 
define probabilistic models
in a similar way to many methods in machine learning and statistics. 
While often SEMs 
will be specified
\textit{by hand} based on 
expert judgement or knowledge, 
in this work we seek to exploit advances in 
probabilistic modeling  
in order to 
infer the structure and form of SEMs directly from observational data.
In particular, we 
focus on 
affine autoregressive flow models \citep{papamakarios2019normalizing}.  
%
We consider the ordering of variables in an affine
autoregressive flow model from a causal perspective, 
and 
show that such models are well suited to performing a variety of 
causal inference tasks. 
Throughout a series of experiments, we demonstrate that autoregressive 
flow models are able to uncover causal structure from purely observational data, i.e., 
\textit{causal discovery}. Furthermore, 
we show that 
when autoregressive flow models are conditioned upon the correct 
causal ordering, they may be  employed to 
accurately answer  interventional and counterfactual queries.

\section{Background}
\label{sec::background}

In this section we 
introduce the class of causal models to be studied and 
highlight their correspondence with autoregressive flow models.

\subsection{Structural equation models}
\label{sec::SEM}

Suppose we observe $d$-dimensional random variables $\textbf{x}= (x_1, \ldots, x_d)$ 
with joint distribution $\mathbb{P}(\textbf{x})$.  
%
A structural equation model (SEM) is here defined 
as a 
collection of $d$ 
structural equations:
\begin{equation}
\label{SEM_eq}
x_j = f_j ( \textbf{pa}_j, n_j), ~~ ~ j=1, \ldots ,d
\end{equation}
together with a joint distribution, $\mathbb{P}(\textbf{n})$, 
over latent disturbance (noise)
variables, $n_j$,  which are assumed to be mutually  independent. 
We write $\textbf{pa}_j$ to denote the parents of the variable $x_j$. 
The causal graph, $\mathcal{G}$, associated with 
a SEM in equation (\ref{SEM_eq})
is a graph
consisting of one node corresponding to each variable $x_j$;
throughout this work we assume $\mathcal{G}$ 
is a directed acyclic graph (DAG). 
It is well known that for such a DAG, there exists a causal ordering (or permutation) $\pi$ of the nodes, such that $\pi(i) < \pi(j)$ if variable $x_i$ is before $x_j$ in the DAG, and therefore a potential parent of $x_j$.
%
%
Thus, given the causal ordering of the associated DAG
%
we may re-write 
equation (\ref{SEM_eq}) as 
\begin{equation}
\label{SEM_eq_causalOrder}
x_j = f_j \left (  \textbf{x}_{ < \pi(j)}  ,  n_j \right ), ~~ ~ j=1, \ldots ,d
\end{equation}
where $\mathbf{x}_{< \pi(j)} = \{x_i: \pi(i) < \pi(j)\}$ denotes all variables before $x_j$ in the causal ordering.  

\subsection{Affine autoregressive flow models}


Normalizing flows seek to express the log-density of observations 
$\textbf{x}\in\mathbb{R}^d$ as an invertible and differentiable 
transformation $T$ of latent variables, $\mathbf{z}\in\mathbb{R}^d$,
which follow a simple base distribution,  $p_{{z}}(\mathbf{z})$. 
The generative model implied under such a framework is:
\begin{align}
\mathbf{z} &\sim p_z (\mathbf{z}), ~~~
\mathbf{x} = T( \mathbf{z})
\end{align}
This allows for the density of $\mathbf{x}$ to be obtained via a change of variables as follows:
\begin{equation}
p_x(\mathbf{x}) = p_{{z}}(\mathbf{z}) | \det J_T(\mathbf{z} )|^{-1}   .
\label{flow_inverse}
\end{equation} 
Throughout this work,  $T$ or $T^{-1}$ will be implemented with neural networks. As such,  an important consideration 
is ensuring the determinant of $T$ can be efficiently calculated. 
\textit{Autoregressive} flow models are designed precisely to achieve this goal by restricting the 
Jacobian of the transformation to be lower triangular \citep{huang2018neural}. 
While autoregressive flows can be implemented in a variety of ways, we consider 
affine transformations of the form:
\begin{equation}
z_j = s_j( \mathbf{x}_{1:j-1} ) + e^{t_j( \mathbf{x}_{1:j-1}  )} \cdot x_j
\label{flow_transformation_zx}
\end{equation}
where both $s_j(\cdot)$ and $t_j(\cdot)$ are parameterized by neural networks \citep{dinh2016density}. 
We write $s_j$, $t_j$ to denote that such neural networks are distinct for each
$j = 1, \ldots, d$. 
Such a transformation can also be trivially inverted as:
\begin{align}
\label{flow_inverse_1}
x_j &= \left (z_j - s_j( \mathbf{x}_{1:j-1} ) \right ) \cdot  e^{-t_j( \mathbf{x}_{1:j-1}  )} 
\end{align}
It is straightforward to extend this last equation 
to the case where the ordering in the autoregressive 
structure of $\textbf{x}$ follows a permutation $\pi$:
\begin{equation}
x_j = - s_j( \mathbf{x}_{<\pi(j)} ) \cdot  e^{-t_j( \mathbf{x}_{<\pi(j)}  )} +  z_j \cdot  e^{-t_j( \mathbf{x}_{<\pi(j)})} 
\label{flow_inverse_2}
\end{equation}


The ideas presented in this extended abstract 
highlight the similarities between 
equations (\ref{SEM_eq_causalOrder}) and (\ref{flow_inverse_2}). 
In particular, both models explicitly define an ordering over variables
and 
both models assume 
latent variables (denoted by $\textbf{n}$ or 
 $\textbf{z}$ respectively)
 follow simple, isotropic distributions. 
%
Throughout the remainder,  we will look to 
build upon these similarities in order to employ
autoregressive flow models for causal inference. 
For the remainder of this extended abstract we write 
$\textbf{n}$ to denote both latent disturbances 
in a SEM and latent variables in an autoregressive flow model. 


\section{Flow-based measures of causal direction}
\label{sec::flowCD}

In this section we exploit the correspondence between nonlinear SEMs 
and 
autoregressive flow models highlighted in 
Section \ref{sec::background} in order to derive new measures of causal direction. 
For simplicity, we restrict ourselves to the case of $d=2$ dimensional data in this section.

\subsection{Autoregressive flow-based  likelihood ratio}
The objective of the 
proposed method is  to uncover the causal direction between two univariate variables 
${x}_1$ and ${x}_2$. 
Denote by $x_1 \rightarrow x_2$ the model where $x_1$ is the parent of $x_2$ in the causal graph (\textit{i.e.} $x_1$ causes $x_2$), for which
the associated SEM is of the form:
\begin{equation}
{x}_1 = f_1( n_1 ) ~~~ \mbox{ and } ~~~ {x}_2 = f_2({x}_1, n_2),
\label{bivariate_eq2}
\end{equation}
where $n_1, n_2$ are 
latent disturbances with factorial joint distributions. 

We follow \citet{Hyvarinen2013} 
and pose causal discovery as a model selection problem. To this end, we seek to compare
two candidate models: $x_1 \rightarrow x_2$ against $x_1 \leftarrow x_2$. 
Likelihood ratios are an attractive way to deciding between alternative models 
and have been proven to be uniformly most powerful when comparing simple hypothesis \citep{neyman1933ix}.
To this end, \citet{Hyvarinen2013} focus on the case of linear SEMs with non-Gaussian latent disturbances and present a variety of methods for estimating the log-likelihood, $\log L_{\pi}$, under each candidate model, where $\pi = (1,2)$ or $\pi=(2,1)$. As such, they compute the log-likelihood ratio as
\begin{equation}
 R = \log L_{1\rightarrow 2 } - \log L_{2 \rightarrow 1 }.
\end{equation} 
The value of $R$ may thus be employed as a measure of causal direction. 
They conclude that $x_1 \rightarrow x_2$ if $R$ is positive and $x_1 \leftarrow x_2$ otherwise. 

In this work, we leverage the expressivity of autoregressive flow architectures in order to 
derive an analogous measure of causal direction for nonlinear SEMs. 
To this end, we note that the 
log-likelihood of the bivariate SEM, $x_1 \rightarrow x_2$, can be computed as:
\begin{align*}
\log L_{1\rightarrow 2 } ( \mathbf{x} ) =& \log p_{x_1}(x_1) + \log p_{x_2|x_1}( x_2 | x_1) \\
=& \log p_{z_1} ( f_1^{-1} (x_1)) + \log p_{z_2} (f_2^{-1}( x_1, x_2)) \\
&+ \log | \det \mathbf{J} \mathbf{f}^{-1}|, 
\end{align*}
where the latter equation 
details how such a
 log-likelihood is computed under an autoregressive flow model. 
 We note that in the context of linear SEMs, as studied by \citet{Hyvarinen2013}, 
 the log determinant term will be equal under both candidate models ($x_1 \rightarrow x_2$ and $x_1 \leftarrow x_2$) and therefore cancel. 

We propose to fit two 
autoregressive flow models, each conditioned on a 
distinct causal order
over variables: $\pi = (1,2)$ or $\pi=(2,1)$. 
For each candidate model we train parameters for each flow via maximum likelihood. 
In order to avoid overfitting we look to evaluate log-likelihood for each model over a 
held out testing dataset. As such, the proposed measure of causal direction is 
defined as:
\begin{align}
\label{eq:flowLR}
\begin{split}
 R = &\log {L_{1\rightarrow 2 }(\mathbf{x}_{test}; \mathbf{x}_{train}, \theta) } \\- &\log { L_{2\rightarrow 1 }(\mathbf{x}_{test}; \mathbf{x}_{train}, \theta') }
\end{split}
\end{align}
where $\log L_{1\rightarrow 2 }(\mathbf{x}_{test}; \mathbf{x}_{train}) $ is the estimated log-likelihood 
evaluated on an unseen test data $\mathbf{x}_{test}$. 
If $R$ is positive we conclude that $x_1$ is the causal variable
and if $R$ is negative we conclude that $x_2$ is the causal variable.
We denote by $\theta$ and $\theta'$ the parameters 
for each flow model respectively. 

\begin{figure*}[ht]
	\begin{center}
		\centerline{\includegraphics[width=1.725\columnwidth]{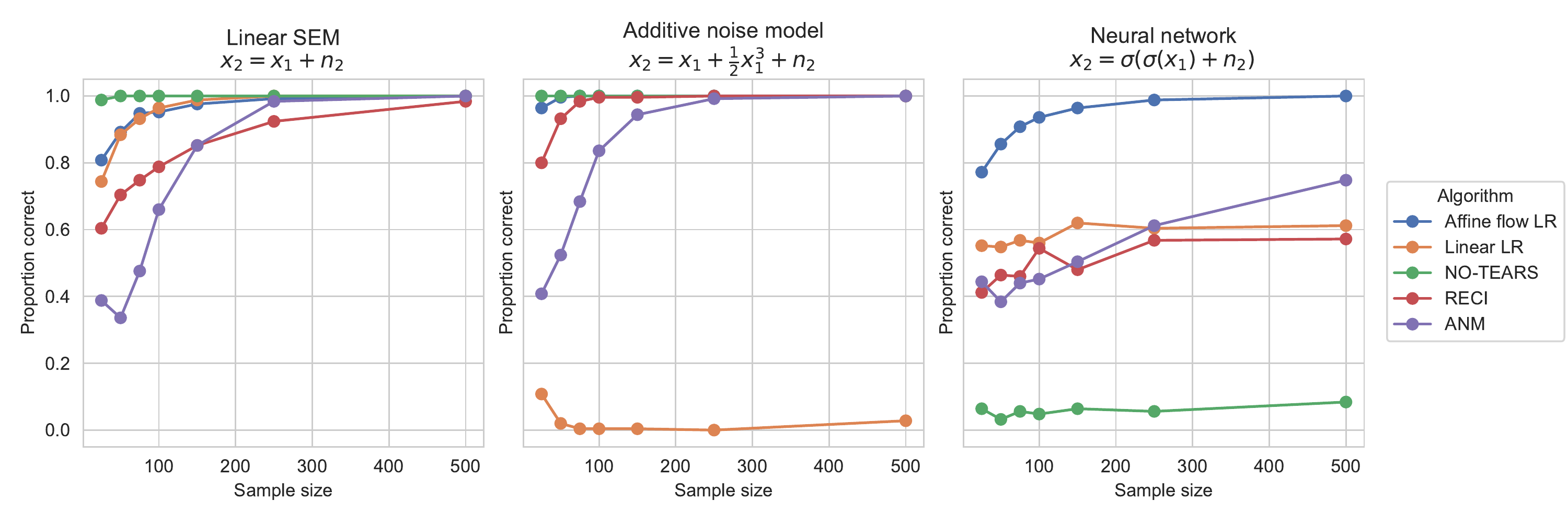}}
		\caption{Performance 
		on synthetic data generated under three
			distinct SEMs. We note that for all three SEMs the proposed likelihood ratio measure of causal discovery
			(Affline flow LR) performs competitively and is able to robustly identify the underlying causal direction. }
		\label{Fig:causalDiscSimulations}
	\end{center}
	\vskip -0.2in
\end{figure*}

\subsection{Experimental results}

In order to demonstrate the capabilities of the proposed method we consider its performance over a variety of synthetic 
datasets as well as on the Cause-Effect Pairs benchmark dataset \citep{mooij2016distinguishing}. 
We compare the performance  against several alternative methods. 
As a comparison against a linear methods we include the 
linear likelihood ratio method of \citet{Hyvarinen2013} as well as the recently
proposed NO-TEARs method of \citet{zheng2018dags}.
We also compare against nonlinear causal discovery methods by considering 
the additive noise model (ANM; \citet{Hoyer2009}). 
Finally, we also
compare against the Regression Error Causal Inference (RECI) method of 
\citet{Blobaum2018}. 
For the proposed flow-based method for causal discovery, we 
employ a two layer autoregressive architecture throughout all synthetic experiments
with a base distribution of isotropic Laplace random variables\footnote{Code to reproduce experiments is 
available at \url{https://github.com/piomonti/AffineFlowCausalInf/}}.


\subsubsection*{Results on synthetic data}
We consider
a series of synthetic experiments where the underlying 
causal model is known. 
Data was generated according to the following 
SEM:
\begin{equation}
x_1 = n_1 ~~~ \mbox{ and } ~~~ x_2 = f( x_1, n_2),\label{eq:SEMgen}
\end{equation}
where $n_1, n_2$ follow a standard Laplace distribution. 
We consider three distinct forms for $f$:
\begin{equation*}
x_2 = f(x_1, n_2) = \begin{cases}
\alpha x_1 + n_2       & \quad \text{linear, } \\
x_1 + \alpha x_1^3 + n_2  & \quad \text{nonlinear, }\\
\sigma \left (  \sigma \left ( \alpha x_1 \right ) + n_2 \right ) &\quad \text{neural net.}
\end{cases}
\end{equation*}
We write $\sigma$ to denote the sigmoid non-linearity. 
For each distinct class of SEMs, we consider the performance of each algorithm under 
various distinct sample sizes ranging from $N=25$ to $N=500$ samples. 
Furthermore, each experiment is repeated 250 times. For each repetition, 
the causal ordering selected at random and 
synthetic data is genererated by 
 first sampling $n_1$ and $n_2$
from a standard Laplace distribution and then passing through equation (\ref{eq:SEMgen}). 


Results are presented in Figure \ref{Fig:causalDiscSimulations}. 
The left panel consider the case of linear SEMs with non-Gaussian disturbances. In such a setting, all
algorithms perform competitively as the sample size increases. 
The middle panel shows results under a nonlinear additive noise model.
We note that the linear likelihood ratio performs  
poorly in this setting. 
Finally, in the right panel we consider a nonlinear model with non-additive noise structure. 
In this setting, only the proposed method is able to consistently uncover the true causal direction. 
We note that the same 
architecture and training parameters were employed throughout all experiments,
highlighting the fact that the 
proposed method is agnostic to the nature of the true 
underlying causal relationship.

\subsubsection*{Results on cause effect pairs data}

We also consider performance of the proposed method on cause-effect pairs benchmark dataset
\citep{mooij2016distinguishing}. 
This benchmark consists of  108 distinct  
bivariate datasets where the objective is to distinguish between cause and effect.
For each dataset,  
two separate autoregressive flow models were trained
conditional on $\pi=(1,2)$ or $\pi=(2,1)$ and
the log-likelihood ratio was evaluated as  in equation 
(\ref{eq:flowLR}) to determine the causal variable.
Results 
are presented in Table 
\ref{sample-table}. We note that the proposed method performs marginally better than alternative 
algorithms.

\begin{table}[h!]
	\caption{Percentage of correct 
		causal variables identified over 108 pairs from the Cause Effect Pairs benchmark.}
	\label{sample-table}
	\vskip 0.15in
	\begin{center}
		\begin{small}
			\begin{sc}
				\begin{tabular}{ccccr}
					\toprule
					Affine Flow LR  & Linear LR & ANM  & RECI    \\
					\midrule
					73 $\%$ & 66$\%$ & 69 $\%$   &  69$\%$ \\
					\bottomrule
				\end{tabular}
			\end{sc}
		\end{small}
	\end{center}
	\vskip -0.1in
\end{table}


\section{Affine flow-based causal inference}
\label{sec::flowCI}

The previous section exploited the fact that autoregressive flows estimate
normalized log-densities of data subject to an ordering over variables.
This allowed for use of likelihood-ratio methods to determine the causal 
ordering over observed variables. 
In this section, we  leverage
the \textit{invertible} nature of flow architectures in 
order to perform both interventional and counterfactual inference. 
We assume that the true causal ordering over variables is known 
(e.g.,  as the
	result of expert judgment or obtained via the methods 
	described in Section \ref{sec::flowCD}).



We now demonstrate how the $do$ operator of \citet{Pearl2009}
can be  incorporated into autoregressive flow models. 
For simplicity, we focus on performing interventions over root nodes in the associated 
DAG, which are assumed known:
this simplifies issues as such nodes have no parents in the DAG and thus there is a one-to-one mapping between the observed variable and 
the corresponding latent variable.\footnote{Performing interventions over non-root variables would require marginalizing over the parents in the DAG.}
As described in \citet{Pearl2009}, 
intervention on a given variable $x_i$ defines a new 
\textit{mutilated} generative model where
the structural equation associated with variable $x_i$ is replaced by the interventional value. 
More formally, the intervention $do( x_i = \alpha)$ changes the structural equation for variable 
$x_i$ from $x_i = f_i( \mathbf{pa}_i, n_i)$ to $x_i = \alpha$. 
This is further simplified if $x_i$ is a root node as this implies that $\mathbf{pa_i} = \emptyset$.\footnote{Changing the structural equation in this fashion for a node that is not a root node implies removing all edges that connects it to its parents, and results in a modified DAG.}
This allows us to directly 
infer the value of the
latent variable, $n_i$, associated with the intervention as
$n_i = f_i^{-1}(  \alpha)$, where $f_i$ is parameterized within the 
autoregressive flow model (see equation \eqref{flow_inverse_2}). Thereafter, we can directly draw samples from 
the base distribution of our flow model for all remaining latent variables and obtain an 
empirical estimate for the interventional distribution by passing
these samples through the flow. This is described in Algorithm \ref{alg:internvention} of the supplementary material.

\subsubsection*{Toy example}
As a simple example we generate  data from the  SEM:
\begin{align}
\begin{split}
\label{intervention_SEM}
x_1&= n_1, ~~~~ x_3 = x_1 + \frac{1}{2} x_2^3 + n_3\\
x_2& = n_2, ~~~~x_4 = \frac{1}{2} x_1^2 - x_2 + n_4
\end{split}
\end{align}
where each $n_i$ is drawn independently from a standard Laplace distribution. 
We consider the expected values of $x_3$ and $x_4$ under various distinct 
interventions to variable $x_1$. 
From the SEM above we can derive the expectations for $x_3$ and $x_4$ under
an intervention  $do(X_1=\alpha)$ as being $\alpha$ and $\frac{1}{2} \alpha^2 $ respectively.

\begin{figure}[!b]
	\begin{center}
		\centerline{\includegraphics[width=\columnwidth]{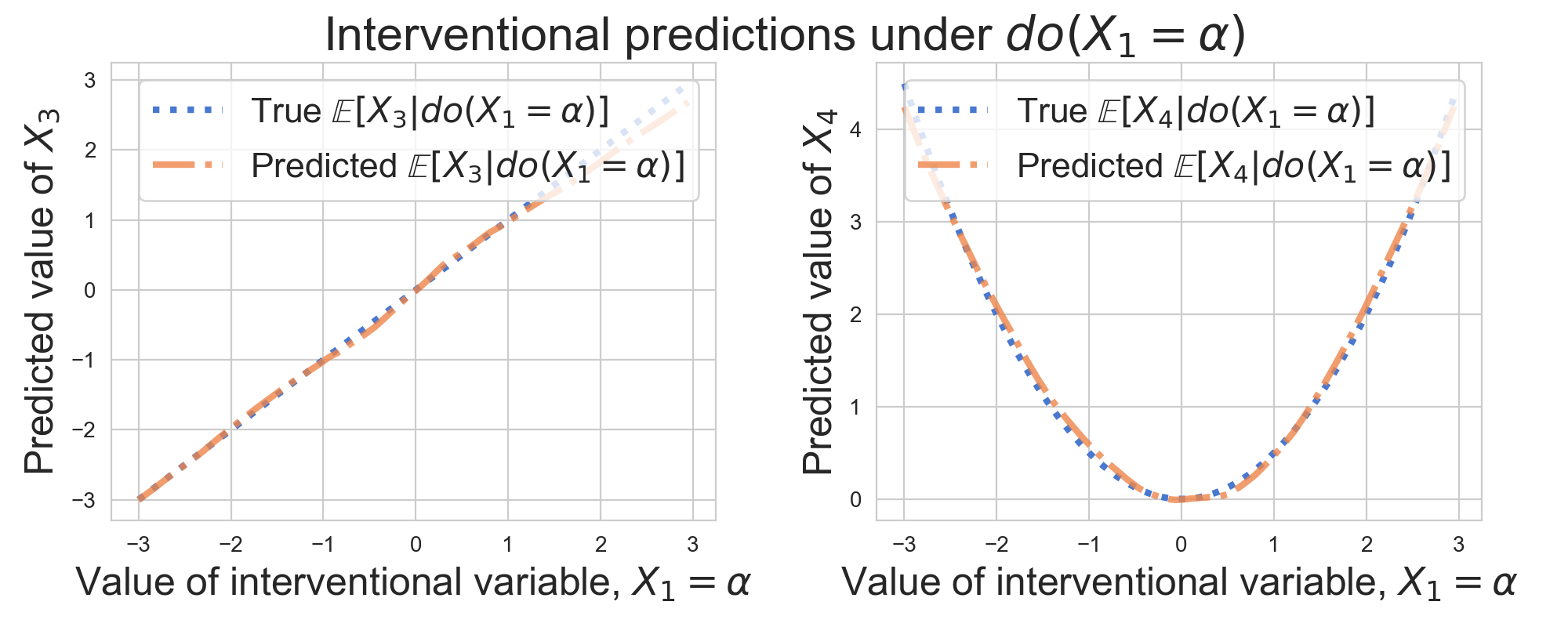}}
		\caption{Interventional predictions for variables $X_3$ and $X_4$ under  $do(X_1=\alpha)$ for $\alpha\in [-3, 3]$. 
		Predicted  expectations obtained
		via an autoregressive flow
		match the true expectations.
			}
		\label{Fig:interventionExample1}
	\end{center}
	\vskip -0.2in
\end{figure}

Figure \ref{Fig:interventionExample1} visualizes the predicted expectations for $x_3$ and $x_4$ under
the intervention $do( X_1=\alpha)$ for the proposed method. We note that 
the proposed autoregressive flow architecture is able to correctly infer the nature of 
the true interventional distributions. 

Furthermore, the invertible nature
of affine flow models also makes then 
suitable to answering 
counterfactual queries.
The fundamental difference between an interventional and counterfactual 
query is that the former seeks to marginalize over latent variables, whereas the latter seeks to infer and 
condition upon latent variables associated with observed data.
In the supplementary material we demonstrate how 
the invertible nature of flow models can be 
exploited to perform accurate counterfactual inference. 


\section{Conclusion}

We argue that
autoregressive flow models are
well-suited to causal inference tasks, ranging from
causal discovery to making interventional 
predictions. 
By interpreting the 
ordering of variables in an autoregressive flow
from a causal perspective 
we are able to learn causal structure by selecting the ordering with the highest test log-likelihood
and  present a
measure of causal direction based on 
the likelihood-ratio for nonlinear SEMs.
We note that 
nonlinear SEMs will
typically not enjoy the same
identifiability guarantees as linear SEMs without
further assumptions  \citep{Hoyer2009, monti2018unified, monti2019causal, khemakhem2020variational},
and in future work we will explore how such assumptions
can be incorporated to flow models. 
Finally, given a causal ordering,
we provide a toy example
showing 
how autoregressive models can be employed to make 
interventional and counterfactual predictions.







\bibliographystyle{plainnat}
\bibliography{library}

\begin{thebibliography}{15}
\providecommand{\natexlab}[1]{#1}
\providecommand{\url}[1]{\texttt{#1}}
\expandafter\ifx\csname urlstyle\endcsname\relax
  \providecommand{\doi}[1]{doi: #1}\else
  \providecommand{\doi}{doi: \begingroup \urlstyle{rm}\Url}\fi

\bibitem[Bl{\"{o}}baum et~al.(2018)Bl{\"{o}}baum, Janzing, Washio, Shimizu, and
  Sch{\"{o}}lkopf]{Blobaum2018}
Patrick Bl{\"{o}}baum, Dominik Janzing, Takashi Washio, Shohei Shimizu, and
  Bernhard Sch{\"{o}}lkopf.
\newblock {Cause-Effect Inference by Comparing Regression Errors}.
\newblock \emph{AISTATS}, 2018.

\bibitem[Dinh et~al.(2016)Dinh, Sohl-Dickstein, and Bengio]{dinh2016density}
Laurent Dinh, Jascha Sohl-Dickstein, and Samy Bengio.
\newblock Density estimation using real nvp.
\newblock \emph{arXiv preprint arXiv:1605.08803}, 2016.

\bibitem[Hoyer et~al.(2009)Hoyer, Janzing, Mooij, Peters, and
  Sch{\"{o}}lkopf]{Hoyer2009}
Patrik~O Hoyer, Dominik Janzing, Joris~M. Mooij, Jonas Peters, and Bernhard
  Sch{\"{o}}lkopf.
\newblock {Nonlinear causal discovery with additive noise models}.
\newblock \emph{Neural Inf. Process. Syst.}, pages 689--696, 2009.

\bibitem[Huang et~al.(2018)Huang, Krueger, Lacoste, and
  Courville]{huang2018neural}
Chin-Wei Huang, David Krueger, Alexandre Lacoste, and Aaron Courville.
\newblock Neural autoregressive flows.
\newblock \emph{International Conference on Machine Learning}, 2018.

\bibitem[Hyv{\"{a}}rinen and Smith(2013)]{Hyvarinen2013}
Aapo Hyv{\"{a}}rinen and Stephen~M Smith.
\newblock {Pairwise Likelihood Ratios for Estimation of Non-Gaussian Structural
  Equation Models}.
\newblock \emph{J. Mach. Learn. Res.}, 14:\penalty0 111--152, 2013.

\bibitem[Khemakhem et~al.(2020)Khemakhem, Kingma, Monti, and
  Hyvarinen]{khemakhem2020variational}
Ilyes Khemakhem, Diederik Kingma, Ricardo Monti, and Aapo Hyvarinen.
\newblock Variational autoencoders and nonlinear {I}{C}{A}: A unifying
  framework.
\newblock In \emph{International Conference on Artificial Intelligence and
  Statistics}, pages 2207--2217, 2020.

\bibitem[Monti and Hyv{\"a}rinen(2018)]{monti2018unified}
Ricardo~Pio Monti and Aapo Hyv{\"a}rinen.
\newblock A unified probabilistic model for learning latent factors and their
  connectivities from high-dimensional data.
\newblock \emph{34th Conference on Uncertainty in Artificial Intelligence
  (UAI)}, 2018.

\bibitem[Monti et~al.(2019)Monti, Zhang, and Hyv{\"a}rinen]{monti2019causal}
Ricardo~Pio Monti, Kun Zhang, and Aapo Hyv{\"a}rinen.
\newblock Causal discovery with general non-linear relationships using
  non-linear {I}{C}{A}.
\newblock \emph{35th Conference on Uncertainty in Artificial Intelligence
  (UAI)}, 2019.

\bibitem[Mooij et~al.(2016)Mooij, Peters, Janzing, Zscheischler, and
  Sch{\"o}lkopf]{mooij2016distinguishing}
Joris~M Mooij, Jonas Peters, Dominik Janzing, Jakob Zscheischler, and Bernhard
  Sch{\"o}lkopf.
\newblock Distinguishing cause from effect using observational data: methods
  and benchmarks.
\newblock \emph{The Journal of Machine Learning Research}, 17\penalty0
  (1):\penalty0 1103--1204, 2016.

\bibitem[Neyman and Pearson(1933)]{neyman1933ix}
Jerzy Neyman and Egon~Sharpe Pearson.
\newblock Ix. on the problem of the most efficient tests of statistical
  hypotheses.
\newblock \emph{Philosophical Transactions of the Royal Society of London.
  Series A}, 231\penalty0 (694-706):\penalty0 289--337, 1933.

\bibitem[Papamakarios et~al.(2019)Papamakarios, Nalisnick, Rezende, Mohamed,
  and Lakshminarayanan]{papamakarios2019normalizing}
George Papamakarios, Eric Nalisnick, Danilo~Jimenez Rezende, Shakir Mohamed,
  and Balaji Lakshminarayanan.
\newblock Normalizing flows for probabilistic modeling and inference.
\newblock \emph{arXiv preprint arXiv:1912.02762}, 2019.

\bibitem[Pearl(2009)]{Pearl2009}
Judea Pearl.
\newblock \emph{{Causality}}.
\newblock Cambridge University Press, 2009.

\bibitem[Pearl et~al.(2009)]{pearl2009causal}
Judea Pearl et~al.
\newblock Causal inference in statistics: An overview.
\newblock \emph{Statistics surveys}, 3:\penalty0 96--146, 2009.

\bibitem[Spirtes et~al.(2000)Spirtes, Glymour, Scheines, Heckerman, Meek, and
  Richardson]{Spirtes2000}
Peter Spirtes, Clark Glymour, Richard Scheines, David Heckerman, Christopher
  Meek, and Thomas Richardson.
\newblock \emph{{Causation, Prediction and Search}}.
\newblock MIT Press, 2000.

\bibitem[Zheng et~al.(2018)Zheng, Aragam, Ravikumar, and Xing]{zheng2018dags}
Xun Zheng, Bryon Aragam, Pradeep~K Ravikumar, and Eric~P Xing.
\newblock {D}{A}{G}s with {N}{O} {T}{E}{A}{R}{S}: Continuous optimization for
  structure learning.
\newblock In \emph{Advances in Neural Information Processing Systems}, pages
  9472--9483, 2018.

\end{thebibliography}


\section*{Supplementary}


\subsection*{Interventions}

Consider an SEM $\mathcal{S} = (\mathbf{S}, \mathbb{P}_\mathbf{n})$, where $\mathbf{S}$ is a set of $d$ equations like in~\eqref{SEM_eq}, and $\mathbb{P}_\mathbf{n}$ is the distribution of the latent disturbances $\mathbf{n}$.
The SEM defines the \emph{observational} distribution of the random vector $\mathbf{x}$: sampling from $\mathbb{P}_\mathbf{x}$ is equivalent to sampling from $\mathbb{P}_\mathbf{n}$ and propagating the samples through the system of equations $\mathbf{S}$.

It is possible to manipulate the SEM $\mathcal{S}$ to create \emph{interventional} distributions over $\textbf{x}$.
This can be done by changing the noise distribution $\mathbb{P}_\mathbf{n}$, or by putting a point mass over one (or many) variable $x_i$, while keeping the rest of the equations fixed --- this latter form was denoted by $\textit{do}(x_i = \alpha)$ in the text above.

Interventions are very useful in understanding causal relationships. If intervening on a variable $x_i$ changes the marginal distribution of another variable $x_j$, then it is very likely that $x_i$ has some causal effect on $x_j$. Conversely, if intervening on $x_j$ doesn't change the marginal distribution of $x_i$, then the latter is not a descendant of $x_j$.

In addition, interventions are used to model the distributions we obtain by running randomized experiments. 
Such experiments are often difficult or unethical to conduct.
Interventional SEMs provide a mathematical framework in which such restrictions are alleviated.

\begin{algorithm}[ht]
	\caption{Generate samples from an interventional distribution}
	\label{alg:internvention}
	\begin{algorithmic}
		\STATE {\bfseries Input:}  interventional (root) variable $x_i$, intervention value $\alpha$, number of samples $S$
		\STATE Infer $n_i$ by inverting flow: $n_i = f_i^{-1}( \alpha)$. 
		\FOR{$s=1$ {\bfseries to} $S$}
		\STATE sample $n_j(s)$ from flow base distribution for $j\neq i$
		\STATE set $n_i(s) = n_i$
		\STATE generate interventional sample as $x(s) = T ( n(s))$, i.e., 
		by passing $z(s)$ through flow. 
		\ENDFOR
		\STATE {\bfseries Return:} samples $\mathbf{x} = \{x(s): s=1, \ldots, S\}$
	\end{algorithmic}
\end{algorithm}

\subsection*{Counterfactuals} 
While 
structural equation models introduce strong assumptions, they also 
facilitate the 
estimation of counterfactual queries.
A counterfactual query seeks to 
quantify statements of the form: what would the value for variable $x_i$ have been if variable $x_j$ had taken 
value $\alpha$, 
 \textbf{given that we have observed} $\mathbf{x}=\mathbf{x}^{obs}$? 
By construction, the value of observed variables 
$\mathbf{x}$ is fully determined by noise/latent variables $\mathbf{z}$ and the associated structural equations, as described in equation (\ref{SEM_eq}). Abusing notation, this may be written as 
$\mathbf{x} = T( \mathbf{n})$ where $T$ encodes the structural equations. 
As such, given a set of structural equations and an observation $\mathbf{x}^{obs}$, 
we follow the notation of \citet{Pearl2009} and write 
${x_i}_{x_j \leftarrow \alpha}(\mathbf{n})$
to denote the 
value of $x_i$ under the counterfactual that $x_j\leftarrow  \alpha$ given observation $\mathbf{x}^{obs}= T(\mathbf{n}^{obs})$.

The fundamental difference between an interventional and counterfactual 
query is that the former seeks to marginalize over latent variables, whereas the latter conditions on the latents. 
The process of obtaining counterfactual predictions is described in 
\citet{pearl2009causal} as consisting of three steps:
\begin{enumerate}
	\item \textbf{Abduction}: given an observation $\mathbf{x}^{obs}$, infer the conditional distribution/values over 
	latent variables $\mathbf{n}^{obs}$. 
	In the context of an autoregressive flow model this is obtained as $\mathbf{n}^{obs}= T^{-1} ( \mathbf{x}^{obs} )$. 
	\item \textbf{Action}: substitute
	the values of $\mathbf{n}^{obs}$ 
	with the 
	values 
	based on the counterfactual query, $\mathbf{x}_{x_j \leftarrow \alpha}$.
	More concretely, for a counterfactual, $\mathbf{x}_{x_j \leftarrow \alpha}$, we 
	replace the 
	structural equations for $x_j$ with 
	$x_j = \alpha$ and 
	adjust the inferred value of latent $n^{obs}_j$ accordingly. 
	As was the case with interventions, 
	if $x_j$ is a root node, then we can set $n^{obs}_j = f_j^{-1}( \alpha )$
	
	\item \textbf{Prediction}:  compute the implied distribution over $\mathbf{x}$ by propagating latent variables, $\mathbf{n}^{obs}$, 
	through the structural equation models. 
\end{enumerate}

\begin{figure}[t!]
	\vskip 0.2in 
	\begin{center}
		\centerline{\includegraphics[width=\columnwidth]{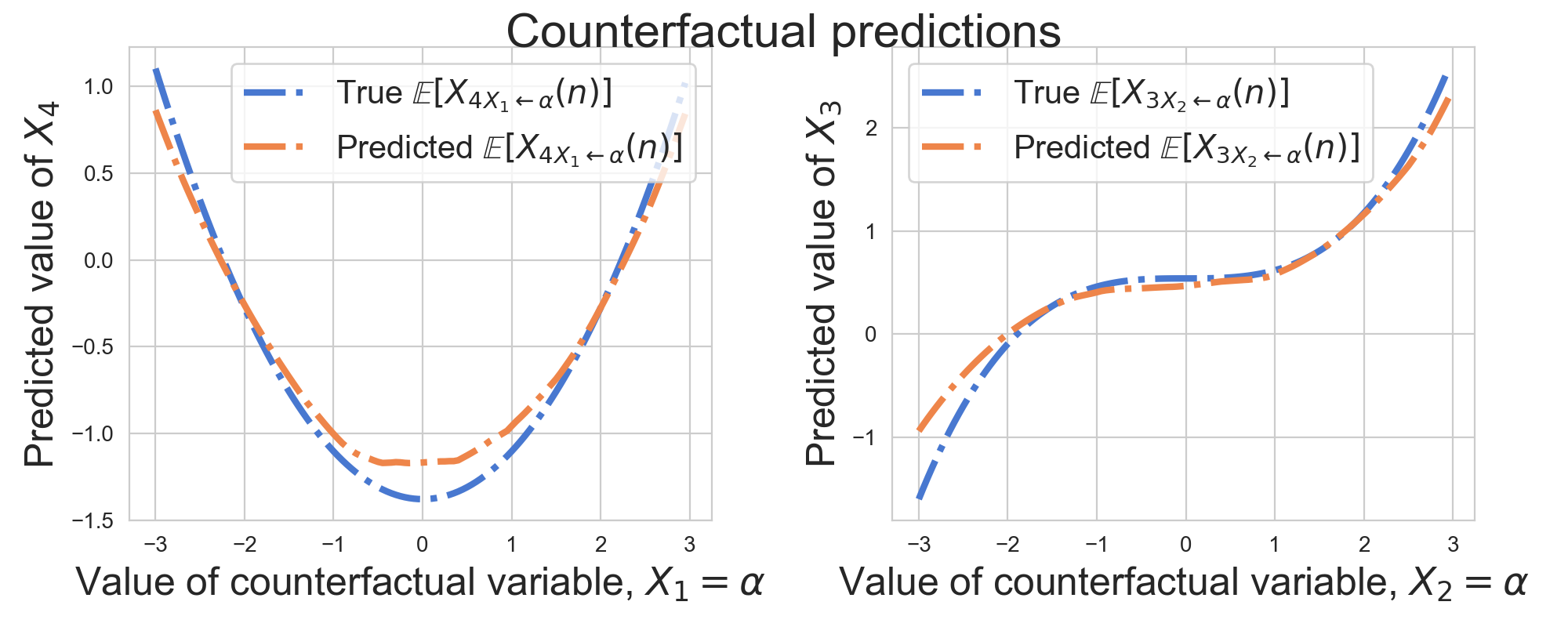}}
		\caption{Counterfactual predictions for variables $x_3$ and $x_4$. Note that 
			flow is able to obtain accurate counterfactual predictions for a 
			range of values of $\alpha \in [-3,3]$.} 
		\label{Fig:counterfactualExample1}
	\end{center}
	\vskip -0.2in
\end{figure}

\subsubsection*{Toy example}
We continue with the simple 4 dimensional structural equation model
described in equation (\ref{intervention_SEM}). 
We assume we observe $\mathbf{x}^{obs} = (2.00  ,  1.50 ,  0.81, -0.28)$
and consider the counterfactual 
values under two distinct scenarios:
\begin{itemize}
	\item  the expected counterfactual value $x_4$ if 
	instead $x_1=\alpha$ for $\alpha \in [-3,3,]$ instead of
	$x_1=2$ as was observed. This is denoted as $\mathbb{E}[{x_4}_{x_1\leftarrow \alpha}(\mathbf{n}) | \mathbf{x}^{obs}]$.
	\item the expected counterfactual value $x_3$ if $x_2=\alpha$ for $\alpha \in [-3,3,]$ instead of
	$X_2=1.5$ as was observed. This is denoted as $\mathbb{E}[{x_3}_{x_2\leftarrow \alpha}(\mathbf{n}) | \mathbf{x}^{obs} ]$.
\end{itemize}
As the true structural equations are provided in equation (\ref{intervention_SEM}), we are able to
compute the true counterfactual expectations and compare these to 
results obtained from an autoregressive flow model. Results, provided in Figure \ref{Fig:counterfactualExample1}, 
demonstrate the the autoregressive flow model 
is able to make accurate counterfactual predictions.

\begin{algorithm}[hb]
	\caption{Generate samples from a counter-factual distribution}
	\label{alg:counterfactual}
	\begin{algorithmic}
		\STATE {\bfseries Input:}  
		observed data $\textbf{x}^{obs}$,
		counterfactual (root) variable $x_i$ and value $\alpha$
		
		\STATE {\bfseries ~~ 1. Abduction}: Infer $\textbf{n}^{obs} = T^{-1} ( \mathbf{x}^{obs})$
		\STATE {\bfseries ~~ 2. Action}: Set $n_j^{obs} = f_j^{-1}( \alpha)$
		\STATE {\bfseries ~~ 3. Prediction}: Pass inferred latent variables forward through flow
		
		{\bfseries Return:} $\mathbf{x}_{x_j \leftarrow \alpha}(\mathbf{n}^{obs}) = T(\mathbf{n}^{obs})$
	\end{algorithmic}
\end{algorithm}
\end{document}